# Assessing learned features of Deep Learning applied to EEG

Dung Truong
SCCN, INC, UCSD, La Jolla CA, USA
dutruong@ucsd.edu
https://orcid.org/

Scott Makeig
SCCN, INC, UCSD, La Jolla CA, USA
smakeig@ucsd.edu
https://orcid.org/0000-0002-9048-8438

Armaud Delorme
SCCN, INC, UCSD, La Jolla CA, USA
CerCo CNRS, Paul Sabatier University,
Toulouse, France
arnodelorme@gmail.com
https://orcid.org/0000-0002-0799-3557

*Abstract*— Convolutional Neural Networks (CNNs) have achieved impressive performance on many computer vision related tasks, such as object detection, image recognition, image retrieval, etc. These achievements benefit from the CNNs' outstanding capability to learn discriminative features with deep layers of neuron structures and iterative training process. This has inspired the EEG research community to adopt CNN in performing EEG classification tasks. However, CNNs learned features are not immediately interpretable, causing a lack of understanding of the CNNs' internal working mechanism. To improve CNN interpretability, CNN visualization methods are applied to translate the internal features into visually perceptible patterns for qualitative analysis of CNN layers. Many CNN visualization methods have been proposed in the Computer Vision literature to interpret the CNN network structure, operation, and semantic concept, yet applications to EEG data analysis have been limited. In this work we use 3 different methods to extract EEG-relevant features from a CNN trained on raw EEG data: optimal samples for each classification category, activation maximization, and reverse convolution. We applied these methods to a high-performing Deep Learning model with state-of-the-art performance for an EEG sex classification task, and show that the model features a difference in the theta frequency band. We show that visualization of a CNN model can reveal interesting EEG results. Using these tools, EEG researchers using Deep Learning can better identify the learned EEG features, possibly identifying new class relevant biomarkers.

## I. INTRODUCTION

Deep learning (DL) is a powerful tool that can extract abstract patterns from complex digital signals without much (if any) feature engineering and can produce impressive classification results in many fields including natural language processing and computer vision [1]. There has been a surge of interest in the EEG research community in applying DL to EEG data. Yet hand-engineered features are still commonly used as inputs in DL-EEG research. Further, beyond achieving classification performance, further uses for DL on EEG data remain unclear [1]. Applying DL directly to raw EEG data not only has the potential to increase classification performance [2], but also allows the possibility of performing automated hypothesis generation by visualizing the trained classification network. Given DL networks' abilities to learn discriminative features between sample classes, the study of these features might help identify unknown brain mechanisms. Furthermore, understanding what features the network learned to base its discriminative judgment on could also aid in optimizing DL

Research supported by NIH grant 1RF1MH125934-01

models trained on EEG data. Many visualization techniques have been applied in the Computer Vision domain to reveal hierarchical feature representation in learned Convolutional Neural Networks (CNNs) [3,4,5]. Despite such success, there has been limited DL-EEG research on identifying physiologically meaningful features that CNNs learn from EEG data. Here we applied three visualization techniques to a CNN model that had shown state-of-the-art performance on a sex-classification task [2].

## II. METHODS

We applied three different methods of visualizing the trained CNN to explore the most relevant EEG features: optimal sample search for each classified category, activation maximization, and reverse convolution. Samples extracted or generated via these methods were analyzed in three ways, based on channel time series, channel spectra, and a dynamic representation (movie) of extracted EEG features.

### A. Visualizing a deep convolutional network

#### a. Best samples for each classification category

One straightforward approach to investigating the categorical decision-making by the network is to see which input sample(s) from the training set gives rise to the highest activations of the network's neurons. Each neuron's activation per input sample can be obtained by passing the sample through the network (forward pass) and retrieving the activation of the neuron of interest. We looked for input samples that maximized the activations of the classification neurons. As our network was trained to perform a binary classification task, we obtained both correctly (true positive and true negative) and incorrectly (false negative and false positive) classified samples, giving us a chance to also examine what features might have fooled the network. The visualization process was as follows:

(1) Run each input in the training set through the network and obtain their summed activation for each classification neuron.

(2) Rank the inputs by the activation level of a selected classification neuron.

(3) Examine the input samples with highest activations for each classification category. For the true positive category, we selected female samples that gave the highest activations in the female-predicting neuron. For true negatives, male samples that induced the highest male-predicting neuron's activations were selected. For false negative, we selected misclassified female samples that gave the highest activations in the male-predicting neuron. For false positives, misclassified male samples with highest activations in the female-predicting neuron were selected.

Since for our dataset, each subject's data contributed about 80 samples to the dataset, we also performed the analysis on the single subject level. To rank the subjects with respect to each classification output neuron, we summed the activation score for each subject over all samples for each class-specific neuron. We then selected 20 subjects with the highest total activations for each category similarly to the process outlined in step (3) in the case of sample selection.

### b. Activation maximization

In contrast to the exhaustive search method above that finds 'best' (most informative) inputs, Activation Maximization (AM) lets the convolutional neural network "imagine" an ideal sample that would maximize a given neural activation [4,9]. AM starts with a random noisy input and updates this input using gradient ascent so as to maximize the activation of the selected neuron.

$$x^* = arg\,max_x(a_{i,l}(\theta, x)) \quad (1)$$

where $x$ is the input, $x^*$ is the optimized input, $\theta$ represents fixed parameters of the trained CNN network, and and $a_{i,l}$ is the activation of $i^{th}$ neuron of the $l^{th}$ layer.

This process can be divided into four steps:

(1) A network neuron is selected.

(2) A sample $x = x_0$ of random values is set as the initial network input.

(3) The gradients with respect to the input $\frac{\partial a_{i,l}(\theta,x)}{\partial x}$ are computed using backpropagation, while the weights and bias parameters of the CNN remain fixed.

(4) Each value of the noise input (here, each scalp channel at each time point, as the input is a channels-by-times array) is updated iteratively to maximize the activation $a_{i,l}$ using the gradients computed at step (3):

$$x \leftarrow x + \eta\frac{\partial a_{i,l}(\theta,x)}{\partial x}$$

where η denotes the gradient ascent step size.

(5) This process terminates when the gradient in (3) reaches a minimum or after a fixed number of iterations. The pattern image $x^*$, the 'preferred' input for this neuron, is then obtained.

This process can be applied to different types of CNN neurons, including the convolutional filters, the neurons in the fully connected layer, and the classification neurons selective to males or females respectively. For neurons in the final layer, we used the unnormalized activations rather than the probability returned by the Softmax procedure. Because Softmax normalizes the final layer output to a probability vector summing to one, maximization of one class probability can be achieved by minimizing the probability of the other class.

***Regularized activation maximization.*** If we let the optimization go unconstrained, research has shown that the optimization will synthesize inputs that favor high frequency patterns hence making them unrealistic, closely resembling the properties of adversarial samples [14]. In the natural image domain, to find a most naturalistic input many regularization techniques have been proposed to constrain the optimization process. The general form of the equation (1) above that introduces regularization is

$$x^* = arg\,max_x(a_{i,l}(\theta, x) - \lambda(x)) \quad (2)$$

where λ is a regularizer that adds constraints to the generated input $x$. Different approaches have been adopted to determine $\lambda(x)$, such as L-p norm decay, blurring, total variation, or normalizing gradients [4,9,15]. These regularization methods can be applied to the AM individually or can be combined together in a cooperative manner. Randomly shifting the input image (jittering) before passing it through the network is a simple yet effective method that can make activation maximization produce sharper natural output images [4]. We experimented with different parameters for the Activation Maximization API provided by a CNN-visualization library [13], then decided to apply total variation and L-1 norm regularization combined with jittering to our network. After 400 iterations, further training did not produce much difference in the output, thereby serving as a natural gradient ascent stopping point.

### c. Reverse convolution

It is possible to map output feature maps of convolutional filters back to the original input space, showing what input pattern originally caused a given activation in the feature maps. This mapping can use a *Deconvolutional Network* (DECONVNET) [5]. A DECONVNET can be thought of as a model that uses the same components (filtering, pooling) but in reverse, so instead of mapping raw data points to a feature space, it propagates the output layer neuronal selectivity back into the original input space. A DECONVNET is attached to each of the layers of its corresponding CNN (CONVNET), providing a continuous path back to sample amplitude. To start, a representative or randomly chosen sample is presented to the CONVNET and activations of neurons are propagated and computed as usual up to the output layer. To examine a given neuron activation, we set all other activations in the same layer to zero and give the activation maps as input to the DECONVNET model. Then we compute the activity of the neurons in the

DECONVNET model by successive (i) deconvolution, (ii) unpooling, and (iii) rectifying. This is then repeated until all input space is reached (the channels by times array that is the output of the DECONVNET).

*Deconvolution:* the term "deconvolution" is actually a misnomer as this is not a deconvolution operation in the mathematical sense of the term, i.e. the mathematical inverse of the convolution operation. Deconvolution only guarantees that the output dimension matches that of the original input. To understand the nature of this 'deconvolution' method, we first need to see how the convolution operation can be represented in matrix form. The convolution of an input $x$ with a kernel/filter $k$, producing a feature map $y$, is

$$y = x * k$$

where $*$ is the convolution operation. This can be also be expressed as a matrix-vector multiplication

$$y' = Wx'$$

where $y'$ is $y$ flattened (the 2-D $y$ matrix flattened as a 1-D vector), $x'$ is $x$ flattened, and $W$ is a sparse matrix whose non-zero elements come from $k$.

The 'deconvolution' operation is taking the transpose of $W$ and multiplying it with $y'$ to obtain

$$z' = W^T y'$$

where $z'$ has the same dimension as $x'$. Once reshaped, $z$ has the same dimension as $x$. As this 'deconvolution' operation is based on the transpose of the kernel matrix $W$, it is also known as "transpose convolution" to distinguish it from mathematical deconvolution as often used in the field of signal processing. Note that the elements in $z$ are not necessarily the same as the elements of $x'$. Only the original dimension is restored. The transposed convolution projects the feature map to a higher-dimensional space, in our case to the dimension of the input of the original convolution, while keeping the connectivity pattern of the convolution.

*Unpooling*: In the CONVET, the max pooling operation is non-invertible. However we can obtain an approximate inverse by recording the locations of the maxima within each pooling region in the CONVET in a set of what's called "switch" variables. In the DECONVNET, the unpooling operation uses these switches to place the reconstructions from the layer above into appropriate locations, preserving the structure of the stimulus [5].

*Rectification*: The CONVNET uses rectified linear units (ReLU) that rectify the feature maps to ensure the feature maps are positive. To obtain valid feature reconstructions at each layer (which should also be positive), we also pass the reconstructed signal through a RELU nonlinearity in the DECONVNET.

The reconstruction obtained from a single activation thus resembles the pattern of the original input signal, with structures weighted according to their contribution toward the feature activation. The visualization process can be described as follows:

(1) Select and propagate a chosen EEG sample through the CONVNET.

(2) Select the target neuron and zero-out the activations of all other neurons in the same layer.

(3) Successively deconvolve, unpool, and rectify the neuron's activation through the DECONVNET until the raw EEG dimension has been reached.

(4) This process is applied repeatedly to all neurons of the DL network to obtain a set of corresponding EEG patterns.

This feature extraction depends on the sample chosen as input. In the example presented below, we used the most informative ('best') samples extracted in section (a), as they were more likely to contain unique features of a given class.

The deconvolution method can only be applied to the network's convolutional layers since it relies on taking the transpose of the convolutional filter. However, a generalization of deconvolution, obtaining the gradient of the neuron's activation with respect to the input sample, allows us to obtain saliency maps showing patterns of input samples contributing to the activation of the neurons in the classifying fully connected layer [9]. This method consists of passing a selected input sample through the network, obtaining the activation score of the classification neuron, then applying backpropagation to obtain the gradient with respect to the input. The magnitude of the gradients at each location in the input sample indicates how important its value is to the neuron's activation. Thus taking the absolute value of these gradients produces a *saliency map* with respect to the target classification neuron [9]. We used the saliency map as a mask to remove EEG inputs that contributed the least to the classification. We linearly interpolated sample values that fell below the 30% quantile (via the *resample* function of Matlab) and visualized these samples as for other methods (see below).

### B. Visualizing optimal input samples

Given the inputs either identified or generated by the methods described in section (A), we used three approaches to visualize their properties.

- We plotted the raw sample time series (channels by times).
- We performed spectral decomposition of the channel data using power spectral density and the Welch method (*spectopo* function of the EEGLAB v2021.1 toolbox) using default parameters. This function returns a graphical depiction of the spectrum. We also plotted scalp topography of the spectrum peaks. Mean spectra averaged across all channels and peaks were identified using the *findpeak* MATLAB function (default parameters). Topographic plotting was performed using the *topoplot* function of EEGLAB.

- We generated an animated scalp activity movie to analyze the EEG dynamics of the optimized samples. The spectral analysis above does not allow us to visualize time series changes. For example if an input sample exhibits frontal theta activity that then appears in the occipital scalp channel region, the mean spectral decomposition would miss such a transition and instead show uniform theta activation in both scalp territories. Movie generation was performed using the *eegmovie* function of EEGLAB v2021.1 [16].

*C. Model used for illustration*

We used a CNN model optimized for biological sex detection that has proven robust when applied to different datasets [2]. The model architecture is typical in the field, and is therefore a suitable example to extract features of interest. Each of the first 4 CNN layers is followed by a max pooling layer then a dropout layer using a 25% dropout rate. The output of each fully connected (FC) convolutional layer (except the last) is transformed by a rectified linear unit (ReLU). The classification layer is a 2-unit FC layer with a softmax activation, resulting in a probability $p$ for male or female sex ($p < 0.5$ for males and $p \geq 0.5$ for female). The number of the network trainable parameters is 12,713,934.

The model was trained on resting data from 1,574 participants in the Healthy Brain Network project archive (50% female) [7]. We split the balanced data into training, validation, and test sets in size ratios 60:30:10. Each segment received a binary label, indicating male (0) or female (1). This gave 71,300 samples (885 participants; 49.94% female) for training, 39,868 samples (492 subjects; 50% female) for validation, and 16,006 samples (197 subjects; 50.3% female) for testing. The model used in this manuscript is the optimized R-SCNN model presented in [2], whose architecture is shown in Table 1.

| Layer | Filter size | # of filters/hidden units |
|---|---|---|
| Convolutional | 3x3 | 100 |
| MaxPooling Dropout (25%) | | |
| Convolutional | 3x3 | 100 |
| MaxPooling Dropout (25%) | | |
| Convolutional | 2x3 | 300 |
| MaxPooling Dropout (25%) | | |
| Convolutional | 1x7 | 300 |
| MaxPooling Dropout (25%) | | |
| Convolutional | 1x3 | 100 |
| Convolutional | 1x3 | 100 |
| Fully connected | | 6144 |
| Fully connected | | 2 |
| Softmax | | |

**Table 1. R-SCNN architecture**. The ReLU activation function processing. The outputs of the convolutional and fully connected layers are not shown for brevity. All convolutional layers have stride 1 and no padding. All pooling layers have window size 2x2, stride 2, and no padding.

The model was trained on a single NVIDIA V100 SMX2 GPU (32 GB) running Python 3.7.10 and PyTorch 1.3.1. During training, the validation data were used to assess its performance and to inform a stopping rule. An Adamax optimizer was used with default hyperparameters (learning rate = 0.002, $\beta_1$ = 0.9, $\beta_2$ = 0.999, $\epsilon$ = 1e-08) except for setting decay = 0.001. Batch size was set at 70, following [2]; training was performed for 70 epochs. Neither other hyperparameter tuning nor batch normalization were performed.

***Model performance.*** Per-sample prediction accuracy was 80.6% (with a 95% confidence interval of 79.7% to 81.5%, obtained by running the model 10 times with different random initializations). Per-subject prediction performance was also assessed, giving an 85.1% accuracy using majority voting out of 40 EEG samples for each subject (with 95% confidence interval of 84.3-85.9%) [2].

### III. RESULTS

*A. Best samples for each category*

As ours was a binary classification task (1 female, 0 male), we retrieved inputs with maximal activation in the classification neurons for four prediction categories: true positive (TP; true positive = female sample predicted as female), true negative (TN; male sample predicted as male), false positive (FP; male sample predicted as female), and false negative (FN; female sample predicted as male).

We propagated each sample in the validation dataset through the trained network to calculate the activations from the two neurons in the last fully-connected layer. We observed that the individual samples that gave rise to highest classification activation for the network (male or female) are not always physiologically plausible samples (Fig. 1). This suggests that some features learned by the network are not easily interpreted by looking at either the time domain or the spectral power domain.

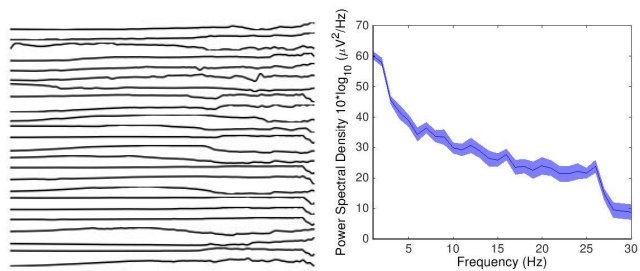

**Fig.1.** Example of non-physiological sample (left) and its spectral decomposition (right) that gave the highest activation for the true positive class (female) the classification neurons. The time series shows strong drifts that likely represent eye movements.

We then summed these activations over subjects (see Methods). For this we selected the top 20 subjects in each of the four classification categories (TP, TN, FP, and FN). Spectral decompositions, averaged across these subjects, showed that TN subjects (male subjects correctly classified) and FN (female classified as male) subjects' data had most similar power spectra. TP (correctly classified female) and FP (male incorrectly classified as female) spectra were closest as compared to the other two categories (Fig. 2, upper panel).

Looking at the spectral decomposition for each subject makes clear the difference between TN and TP, indicating that these subjects are easy to classify by the network. Even though the difference is maximum in the alpha frequency range (near 10 Hz), the network takes advantage of differences in all frequency bands. The difference between error categories FN and FP is mostly constant across the 1-30 Hz frequency, although the difference is maximal in the theta (4-8 Hz) and high beta (20-23 Hz) (Fig. 2, lower panel).

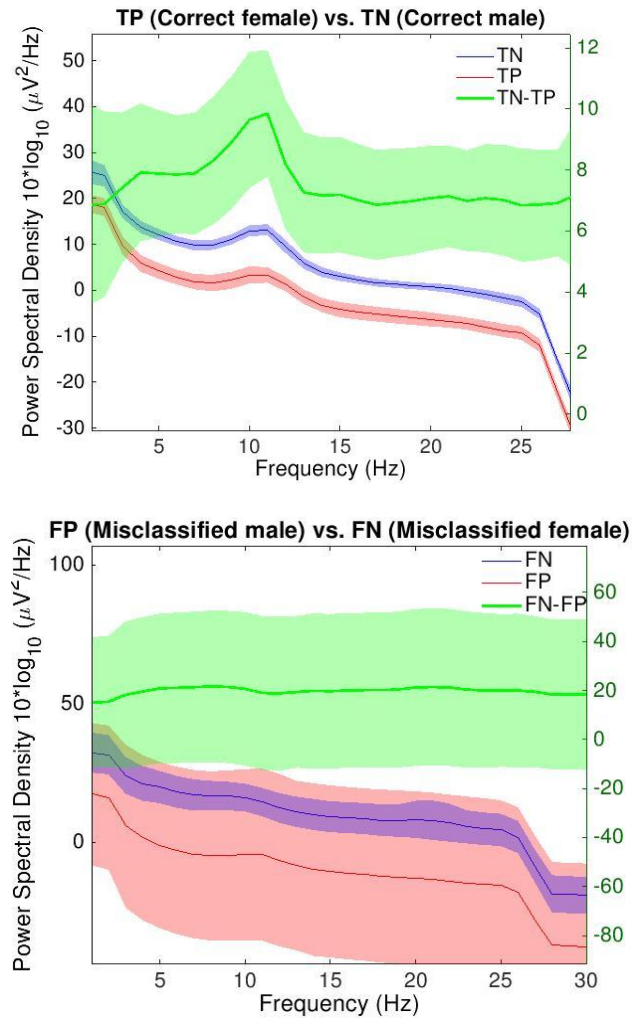

**Fig.2.** Channel- and subject-mean spectral decompositions for samples that produced maximum neuron activation in the last fully connected layer for four classification categories, with 95% confidence interval out of 20 subjects. Top: TN (blue), TP (red) and difference between the two (green). Bottom: FP (blue) and FN (red) and difference between the two (green). Shaded areas indicate 95% confidence intervals calculated across the selected participants.

### B. Activation maximization

***Convolutional layers.*** Spectral decomposition on synthesized inputs of the convolutional filters for all convolutional layers showed that the convolution were selective to a broad range of frequencies (Fig. 3).

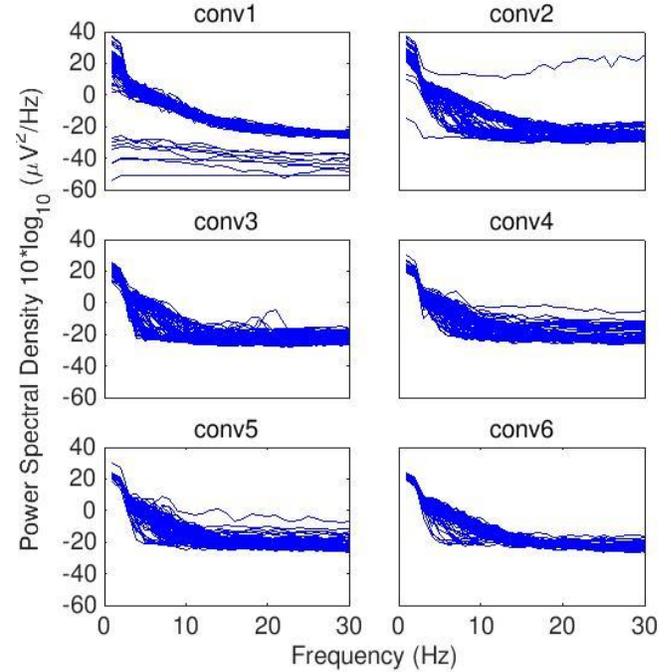

**Fig. 3.** Power spectra of the synthesized raw EEG samples that maximize the activations of 100 filters for each of the convolutional layers (see Methods) from conv1 to conv6 (left to right, top to bottom).

Power spectra of raw EEG samples maximizing activation of the first convolutional layer did not show any distinct frequency selectivity and exhibited a clean inverse spectral curve typical of raw EEG (Fig. 3). Deeper in the network, the filter generated inputs showed contribution in the theta and alpha power band. We also observed that some filters (in layer 4 and 5) had different patterns of selectivity to information from frequencies above 25 Hz. They might be using this as feature information, or to flag some trials as artifacts.

***Last fully connected layer.*** Synthesized raw EEG input that would maximize the activations of the two classification neurons of the last fully connected layer, before softmax regularization, could be seen as samples that the two neurons "imagined" as most representative of the male and female classes (Fig. 4).

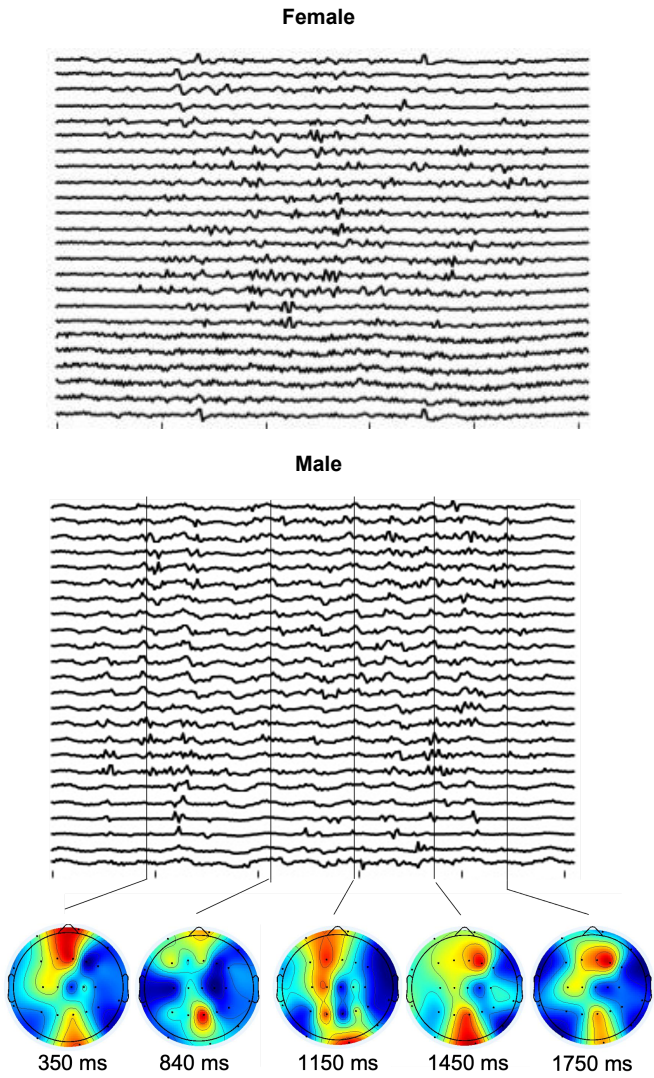

**Fig. 4.** Female (top) and male (bottom) data samples synthesized by maximizing the activation of the classification neurons in the last fully connected layer before softmax calculation. For illustration purposes, the bottom row shows snapshots at different latencies of the scalp topography movie.

Spectral decomposition of the two synthetized samples showed a distinct peak in the 7-8 Hz high theta power range for males but not for females (Fig. 5). To account for the variability associated with random initialization, we synthesized 20 different male and female samples.

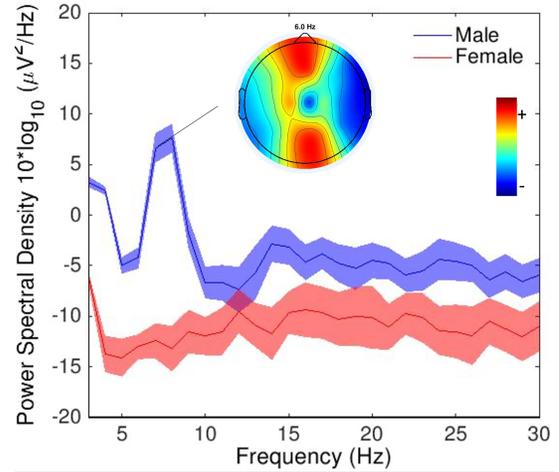

**Fig.5.** Spectral decomposition for 20 data samples synthesized so as to maximize the activations of the male-selective classification neurons (blue) and the female selective one (red). Shaded regions indicate 95% confidence intervals. Scalp power topography for 1 randomly selected sample out of the 20 synthesized ones in Fig. 4 at 6 Hz is also plotted.

### C.  Reverse convolution

***Convolutional layers.*** We observed that for some convolutional layers, not all filters were active when passing the samples through the network. The last convolutional layer (conv6) in particular had very few of the 100 available filters activated. We also observed different patterns of activation for the convolutional layers as compared to the previous method of maximal activation. The network tends to reconstruct inputs with more uniform frequency distribution, seemingly weighting all frequencies similarly.

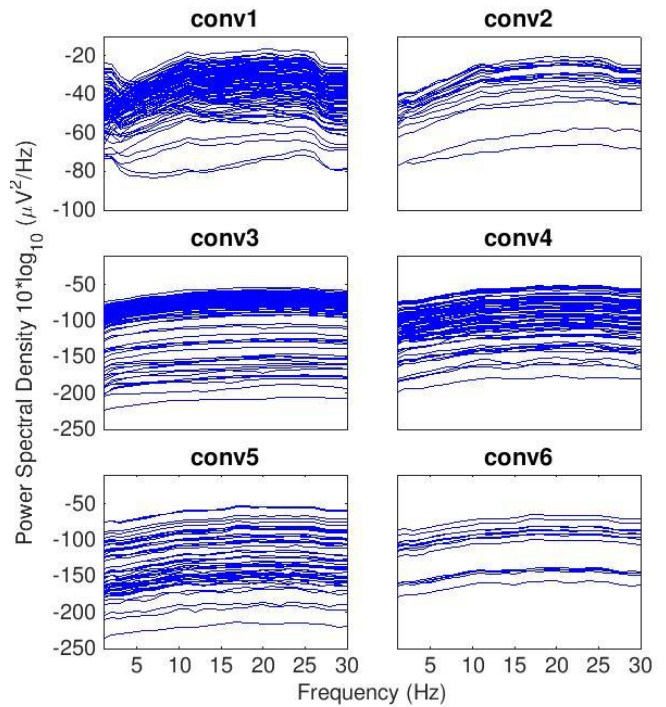

**Fig.6.** Spectra decomposition from 1-30 Hz (x-axis) for active filters of the convolutional layers conv1 to conv6 (left to right, top to bottom). Spectra are constructed by averaging the deconvolution of 10 randomly selected samples across all channels (see Methods).

***Saliency map for classification layer.*** For classification neurons, we can not deconvolve input and can only build saliency maps (see Methods). Fig. 7 shows the saliency map for a female sample with respect to the female classification neuron. Saliency maps are used to mask input samples and emphasize learned features in these samples. We observed that for a large number of samples, the theta and alpha bands were used as a feature. We also observed that for all samples, the saliency maps with respect to both male and female classification neurons emphasized the same frequency bands, indicating that for a given sample, the same frequencies provide class-discriminating information and hence similarly emphasized.

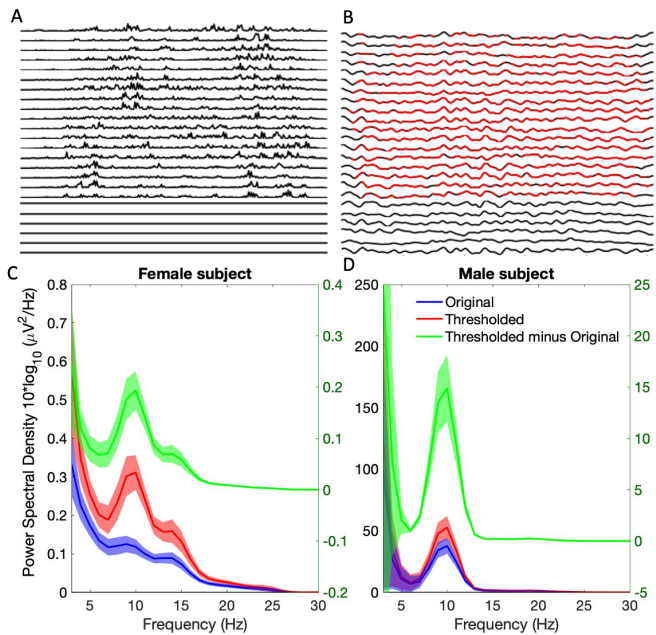

**Fig.7.** Deconvolution/saliency of classification layer. A. Example of a saliency map for a 24-channel 2-second female sample (see Methods). B. Original sample with saliency masking at 30% quantile threshold in red (see Methods). C. Averaged power spectrum of 35 original and thresholded samples of a female subject and 95% confidence interval in shaded area. D. Same as C for samples of a male subject.

D. Hypothesis validation

Finally, we computed the spectral difference between male and female samples for the validation set (Fig. 8). We observed differences in the high theta 5-8 Hz band and in the alpha band, validating previous approaches. Differences were also observed in the alpha frequency band.

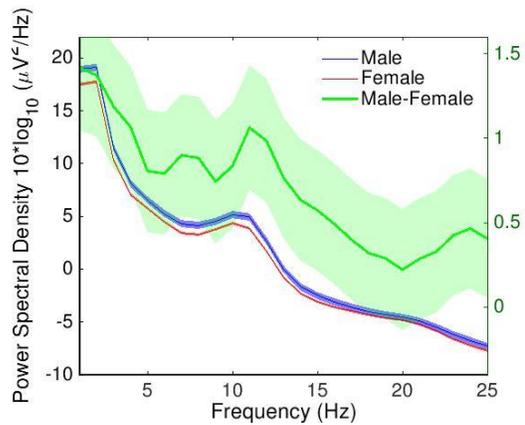

**Fig. 8.** Power spectra of male and female samples from validation set with 95% confidence interval (left axis), and mean spectral power difference between male and female (right axis).

IV. DISCUSSION

Here we showed using three methods how to visualize learned features of a DL binary classification network applied to raw EEG data samples. Our visualization result for the convolutional layer showed more richer frequency features as the depth increase (Fig. 3), similar to previous work on sex classification using DL EEG [6]. Although differences were present between the two classes in both the alpha and high theta frequency bands, the network seemed to mostly focus on features in the theta frequency band. This is because differences in the alpha frequency band are likely due to samples easy to classify where differences in other frequency bands are also present (Fig 2 top). However, for samples difficult to classify (Fig 3 bottom), theta is a more powerful feature.

Another promising approach to generating optimal data samples for the network, not attempted here, would use an Generative Adversarial Network trained to generate realistic samples given the training distribution. Unlike activation maximization, which is constrained by a regularization method, GAN-AM [17] does not impose any prior constraints, thus might generate more optimized and realistic EEG samples.